# Learning Bayesian Networks from Incomplete Data with Stochastic Search Algorithms

**James W. Myers**
George Mason University
Fairfax, VA 22032-4444
myers29@erols.com

**Kathryn Blackmond Laskey**
George Mason University
Fairfax, VA 22032-4444
klaskey@gmu.edu

**Tod Levitt**
IET
Setauket, NY 11733
tlevitt@iet.com

**Abstract**

This paper describes stochastic search approaches, including a new stochastic algorithm and an adaptive mutation operator, for learning Bayesian networks from incomplete data. This problem is characterized by a huge solution space with a highly multimodal landscape. State-of-the-art approaches all involve using deterministic approaches such as the expectation-maximization algorithm. These approaches are guaranteed to find local maxima, but do not explore the landscape for other modes. Our approach evolves structure and the missing data. We compare our stochastic algorithms and show they all produce accurate results.

## 1  INTRODUCTION

Bayesian networks are growing in popularity as the model of choice of many AI researchers for problems involving reasoning under uncertainty. They have been implemented in applications in areas such as medical diagnostics, classification systems, software agents for personal assistants, multisensor fusion, and legal analysis of trials. Until recently, the standard approach to constructing belief networks was a labor-intensive process of eliciting knowledge from experts. Methods for capturing available data to construct Bayesian networks or to refine an expert-provided network promise to greatly improve both the efficiency of knowledge engineering and the accuracy of the models. For this reason, learning Bayesian networks from data has become an increasingly active area of research. Most of the research to date has relied on the assumption that data are complete; that is, the values of all variables are known for all cases in the database. This assumption is not very realistic since most real world situations involve incomplete information.

Learning a Bayesian network can be decomposed into the problem of learning the graph structure and learning the parameters. The first attempts at treating incomplete data involved learning the parameters of a fixed network structure [Lauritzen 1995]. Very recently, researchers have begun to tackle the problem of learning the structure of the network from incomplete data. A major stumbling block in this research is that when information is missing, closed form expressions do not exist for the scoring metric used to evaluate the network structures. This has led many researchers down the path of estimating the score using parametric approaches such as the expectation-maximization (EM) algorithm [Dempster, Laird et al. 1977], [Friedman 1998]. The EM algorithm is a proven approach for dealing with incomplete information when building statistical models [Little and Rubin 1987]. EM and related algorithms show promise. However, it has been noted [Friedman 1998] that the search space is large and multimodal, and deterministic search algorithms are prone to find local optima. Multiple restarts have been suggested as a way to deal with this problem.

An obvious choice to combat the problem of "getting stuck" on local maxima is to use a stochastic search method. This paper explores the use of evolutionary algorithms (EA) and Markov chain Monte Carlo (MCMC) algorithms for learning Bayesian networks from incomplete data. We also introduce an algorithm, the Evolutionary Markov Chain Monte Carlo (EMCMC) algorithm, which combines the advantages of the EA and MCMC, which we believe, advances the state of the art for both EA and MCMC. We will also introduce an adaptive mutation approach for proposing new states for MCMC that can be thought of as a meta-MCMC.

In addition to a robust means for learning Bayesian networks from incomplete data, to the best of our knowledge, this is the first study to compare the performance of evolutionary algorithms and Markov Chain Monte Carlo algorithms. While the goals of each of these algorithms are slightly different they are both effective approaches to learning.

We'll begin by describing the EA and MCMC algorithms. Since these algorithms evolved from different fields, we will attempt to clarify terms with common meaning. We will discuss the common representation used and the common fitness function. In section 3 we will introduce the EMCMC and describe why it is both an EA and a MCMC. We describe what advantages it promises and give the algorithm. We also discuss the adaptive mutation operator and describe why it can be thought of as a meta-MCMC. Section 4 describes our empirical approach and provides results. We conclude, in Section 5, with a summary and direction for future research.

## 2  BACKGROUND

### 2.1  EVOLUTIONARY ALGORITHM

Evolutionary algorithms are a family of algorithms modeled after the organic evolutionary processes found in nature. They consist of a population of individual solutions that are selected and modified in order to discover overall better solutions in the search space. More specifically, the algorithms proceed as follows. An initial population of solutions is generated. Until some stopping criterion is met the population is evolved in the following manner. Individuals are selected from the population based on fitness. The fitness of an individual is determined by how good a solution it provides. The selected individuals are then modified using genetic operators. The most common genetic operators are crossover and mutation. In crossover, two parent individuals are selected and information is exchanged between the individuals at selected points, see Figure 1. In mutation, a single individual is modified in some way, such as flipping a bit in a binary string representation. The resulting individuals are known as the offspring. The final step of the evolutionary algorithm is to select the next generation from the current parent population and the offspring population. Each new generation follows this process until the stopping criterion is met.

As was noted in Section 1, there is no closed form expression for evaluating structures when the data are incomplete. One approach is to turn the incomplete data problem into a complete data problem by evolving the missing data and imputing these values into the data. This allows us to use the Bayesian Dirichlet scoring metric, BDe, developed by Cooper and Herskovits and Heckerman et al., [Cooper and Herskovits 1992], [Heckerman, Geiger et al. 1995].

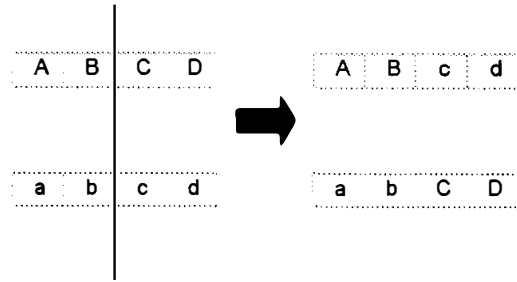

**Figure 1 Crossover Operator**

By imputing samples into the data, the search space becomes more complex. We now must search over the missing data and network structures. We take the unique approach that evolves both the missing values (samples) and the structures simultaneously [Myers, Laskey et al. 1999]. This approach requires that we define representation for both the missing data and the structures. The missing data representation is straightforward. We represent each cell from the dataset that has a missing value as a gene. The gene takes on sampled values from the set of values of the corresponding variable. The chromosome is a string of missing values.

The structure, $B_S$, can be represented as an adjacency list, see Figure 2, where each row represents a variable $V_i$, and the members of each row, with the exception of the first member are the parents of $V_i$, $pa(V_i)$. The first member of each row, i.e. the first column of the adjacency list, is the variable $V_i$. Although we show it in the picture for clarity, the internal representation encodes the parents only, with the variable being encoded by position. The adjacency list can be thought of as a chromosome, where each row is a gene and the

pa($V_1$) are the alleles. This representation is convenient because the log form of BDe is the summation of scores for each variable. Because of this, each gene can be scored separately and added to generate the fitness score for the entire structure.

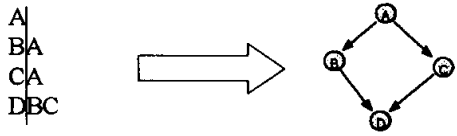

**Figure 2 Structure Mapping Genotype to Phenotype**

The allele values that each gene can take on can become enormous. The values can range from no parents to n-1 parents, where n is the number of variables in the dataset. Thus an allele can take on $\sum_{i=1}^{m} \binom{n-1}{i}$ possible values where m is the maximum set of parents a variable can have and n is the number of variables in the dataset. As an example of the large size of allele values take n=11 and m=4, the number of possible values for a given allele is 376, while if n=41 and m=4, the number grows to 102,091.

In addition to the large combination of allele values per gene, the genes are highly correlated. This is because the alleles are combinations of other genes as parents. Many combinations can lead to illegal structures; in other words, structures that are not directed acyclic graphs. This problem is alleviated by arbitrarily assigning illegal structures a very low score. The reason for allowing illegal structures is the chromosome may contain very good genes and if selected as parents the genes can be reconstituted as building blocks for even better structures through recombination or mutation

The genetic operators require some explanation. For the missing data and structure chromosomes we chose uniform parameterized crossover [Syswerda 1989], [DeJong and Spears 1990]. Uniform parameterized crossover selects each gene for crossover probabilistically. If the parameter is set to 0.5, then on average about half the genes will be crossed over. Figure 3 depicts uniform crossover for the structure chromosome.

Mutation for the missing data chromosome is accomplished by randomly selecting a gene and then randomly selecting from the remaining values of the corresponding variable. For the structure chromosome the mutation operator is tailored to the representation we used and its mapping to a directed graph phenotype. Recall the gene of the structure chromosome represents the gene's parent nodes in the graph. We include two basic modifications to a gene: add a node and delete a node. These operators have the effect in the phenotype of adding and deleting arcs, respectively. We also include a third basic modification, reversal of an arc, which is implemented genotypically by deleting the parent-child arc and adding a child-parent arc.

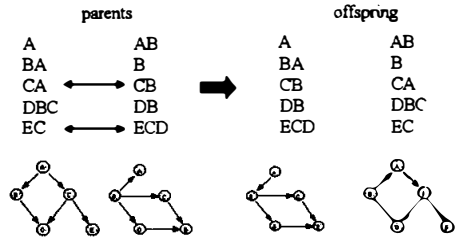

**Figure 3 Uniform Crossover**

## 2.2 MARKOV CHAIN MONTE CARLO ALGORITHM

Markov Chain Monte Carlo is a general class of algorithms used for optimization, search, and learning. These algorithms arose in statistical physics where they were used as models of physical systems that seek a state of minimal free energy. Markov Chain Monte Carlo algorithms have also been used more recently in statistical inference and artificial intelligence [Neal 1993], [Gilks, Richardson et al. 1996], [Geman and Geman 1984].

Statistical physicists model physical systems in terms of their macrostates and microstates. A macrostate is a system's observable components, such as temperature and pressure. A microstate is the non-observable detailed state of the system's atomic structure, such as position and velocity of every molecule in the system.

Each microstate has an associated energy, E(s). An isolated system free of external influences is assumed to evolve to an equilibrium state that minimizes a quantity known as *free energy*. This equation is probabilistic—that is, it is not possible to know the microstates in detail, but only to predict the probabilities of various microstates. The equilibrium probability that a system is in a given microstate is

$$P(s) = \frac{1}{Z}\exp\left(\frac{-E(s)}{T}\right) \quad (1)$$

where Z is a normalization constant which ensures that the probabilities sum to 1, and T is the temperature. This distribution is commonly known as the Boltzman distribution. The Boltzman distribution is the probability distribution over microstate that minimizes free energy, subject to a constraint on the expected value of the total energy of the system. Free energy is defined as F=<E>-TS, where <E> is the expected value of the energy, T is temperature, and S is the entropy. MCMC algorithms for optimization and statistical inference involve creating a mapping from the minimal free energy distribution (1) to the solution of the optimization or inference problem.

Statisticians recognized that any probabilistic inference problem can be translated into the language of statistical physics. The probability distribution of any random variable can by represented as a Boltzman distribution over the microstates of an imaginary physical system. If S is the sample space and P(s) is the probability of occurrence of s ∈ S, then we define an associated energy as

$$E(s) = -T[\log P(s) + Z] \quad (2)$$

where T>0 is an arbitrary constant and Z is chosen as the solution of the equation

$$Z = \sum_s \exp\{-T[\log P(s) + Z]\} \quad (3)$$

It is clear that P(s) is the Boltzman distribution of the system with energy values E(s) and temperature T. Note that when T is set to 1, then E(s)=-logP(s).

Consider a generic problem of infering the posterior distribution of a parameter θ from a set of observations $x_1,...,x_C$, where the observations are assumed to be independent samples from the distribution P(x|θ). The posterior distribution can be defined in terms of a Boltzman distribution as

$$P(\theta|x_1,...,x_C) = \frac{1}{Z_C}\exp(-E(\theta)) \quad (4)$$

The representation of (4) for a general statistical inference problem means that algorithms from statistical physics can be applied to general problems in statistical inference. The application of methods from statistical physics to complex problems in statistical inference is a burgeoning area of research. Many algorithms from statistical physics, both deterministic and stochastic, are becoming popular in statistics. Here we focus on a class of algorithms called *Markov Chain Monte Carlo*.

A first order Markov chain is a series of random variables, $X^0, X^1, ..., X^n$, where the probability distribution for $X^n$ depends on $X^0, X^1,...,X^{n-1}$ only through the value $X^{n-1}$. That is, $X^n$ is independent of the past history given $X^{n-1}$. Markov chains can be described in terms of transition probabilities and states. In other words, the probability of a Markov chain being in state x at time n+1 is given by

$$P_{n+1}(x) = \sum_{x'} P_n(x')T_n(x',x) \quad (5)$$

where $T_n(x',x)$ is the transition matrix that defines the probability of moving from state x' to x. The distribution in (5) is said to be stationary if it persists forever once reached. If π(x) is a stationary distribution, then (5) can be rewritten as

$$\pi(x) = \sum_{x'} \pi(x')T_n(x',x) \quad (6)$$

If a Markov chain satisfies certain regularity conditions [Feller 1968], then it converges to a unique stationary distribution. We can construct a Markov chain with a specified Boltzman distribution as its stationary distribution by ensuring that the transition probabilities satisfy a condition known as detailed balance [Neal 1993], also known as local reversibility. Detailed balance ensures that at equilibrium, transitions from any state x to any other state x' are balanced probabilistically by transitions from x' back to x. More formally, a Markov chain satisfies detailed balance if

$$\pi(x)T(x,x') = \pi(x')T(x',x) \quad (7)$$

It is straightforward to verify that a chain satisfying (7) has stationary distribution π(x). Additional conditions are required on the transition probabilities to ensure that the chain converges to the distribution π(s) from any initial distribution [Feller 1968]. Detailed balance is a stronger condition than necessary for convergence to a stationary distribution. That is, a Markov chain may converge to a stationary distribution without detailed balance holding. However, detailed balance gives a simple recipe for designing algorithms that converge to a stationary distribution specified up to a normalization constant (i.e., a Boltzman distribution).

There are several common ways to construct a sampler that satisfies detailed balance. We applied one of the most common sampling approaches, known as Metropolis-Hastings sampling [Metropolis, Rosenbluth et al. 1953], [Hastings 1970]. The Metropolis-Hastings algorithm samples from a joint distribution by repeatedly generating random changes to the variables and then accepting or rejecting the changes in a way that preserves detailed balance. In this case the transition probabilities of the Markov chains consist of a proposal distribution and acceptance probability

$$T(x,x') = S(x,x')A(x,x') \qquad (8)$$

where $S(x,x')$ is the proposal distribution and $A(x,x')$ is the acceptance probability.

The proposal distribution is used for generating the next candidate state, $x^*$. It can be as simple as adding a sample from a Gaussian to the current real-valued state or as complex as randomly adding or deleting arcs in a Bayesian network structure. After generating the candidate state, $x^*$, the state is evaluated and accepted probabilistically. The acceptance distribution is given by

$$A(x,x') = \min\left(1, \frac{P(x')}{P(x)} \cdot \frac{S(x',x)}{S(x,x')}\right) \qquad (9)$$

where for the Boltzman distribution

$$\frac{P(x')}{P(x)} = \exp\left(-\frac{E(x') - E(x)}{T}\right) \qquad (10)$$

The new state x' is accepted with probability $A(x,x')$. It is a simple matter to show that detailed balance holds for the Metropolis-Hastings algorithm, see [Neal 1993].

A major advantage of the MCMC approach is that at stationarity, the Markov chain generates samples from the posterior distribution of interest. That is, the long-run frequency with which a structure is visited is equal to the posterior probability of that structure. Thus, if the algorithm is run sufficiently long, the samples generated can be used to estimate statistics from the posterior distribution such as modes, mean, variance, etc. A main drawback of MCMC is that convergence can be slow and difficult to recognize. A chief reason for slow convergence, and for the inability to recognize when convergence has been achieved, is slow mixing. A Markov chain is said to "mix well" when it moves rapidly through the state space, traversing all regions of the state space in a short time. A chain that mixes well will converge rapidly, and it will not take long to obtain samples that can be treated as independent realizations of the distribution of interest.

Our approach to learning Bayesian networks from incomplete data is to set up two sets of Markov chains for sampling from the incomplete data and the network structures [Myers 1999]. This is synonymous with the two populations in our evolutionary algorithms. The energy term in (10), $E(x)$, is the Bayesian Dirichlet score.

The missing data and network structures have the same representation as given for the evolutionary algorithm. The proposal distributions are both equivalent to the mutation operators from the evolutionary algorithms. Changes are made to the missing data by choosing cells to modify from the string of missing data and then randomly assigning a value from the set of values (less the current value of the cell) for the variable the missing value was originally assigned in the dataset. A new state is proposed for the structure by adding, deleting, or reversing the arcs in the structure. The Metropolis-Hastings criterion (9) is used to determine whether changes are accepted, thus ensuring that the chain satisfies detailed balance. In order to compare with the two population-based algorithms, we ran a population of independent chains in parallel.

## 3 EVOLUTIONARY MARKOV CHAIN MONTE CARLO

### 3.1 WHY EMCMC

Evolutionary algorithms can be characterized as sampling from modes of the distribution of interest. They use selection pressure to balance exploration and exploitation. By selecting and exchanging information between better-fit individuals, the result is an overall better fit population. Evolutionary algorithms have been shown through many empirical studies to work very well for many problems. A problem with evolutionary algorithms is there is little theory to predict their long-term behavior for any particular application. In addition, without prior experience and expectations within a domain, it may be difficult to determine if the solution discovered is acceptable. An additional problem many evolutionary algorithms have is genetic drift, where the majority of the population drifts to a single mode and search essentially stops.

Markov Chain Monte Carlo algorithms evolve to sample from the target stationary distribution. This distribution is equivalent, in statistical physics terms, to the minimum free energy of an imaginary physical system. Unlike the EA we can use the first principles of probability theory to predict the long-term behavior of MCMC. Unfortunately, a MCMC may take a while to converge to a stationary distribution, due to slow mixing.

Given the ability of the EA to exchange information in order to improve fitness, it seems reasonable to conjecture that taking the same approach with MCMC will speed convergence by finding better fit solutions faster. See Holmes and Mallick for a similar approach [Holmes and Mallick 1998]. For the EA, a MCMC approach may also lead to a more diverse population after convergence. This is because the MCMC is sampling from the stationary distribution and not a single mode. In addition, an MCMC based EA samples directly from the posterior distribution of interest allowing us to make theoretically sound statements about the landscape and solutions discovered in the landscape.

### 3.2 THE ALGORITHM

Our approach is to combine the information exchange (crossover) benefits of the canonical evolutionary algorithm with the MCMC while maintaining detailed balance. We call this new algorithm the Evolutionary Markov Chain Monte Carlo (EMCMC) algorithm. The EMCMC is both an EA, in that it is a population-based algorithm with genetic operators such as crossover and mutation, and a MCMC because it is a population of Markov chains that are evolved using the Metropolis-Hastings algorithm.

The first step in the algorithm is to generate an initial population of missing data and structures. The fitness of the structure is determined by imputing the missing data into the dataset and using the Bayesian Dirichlet score to calculate its log probability. The algorithm then iterates through the following steps until a stopping criterion is reached. Two individuals in the population are selected at random. Then, also at random, a decision is made whether to mutate or exchange information via crossover. If mutation is selected, each individual is mutated as described in Section 2.2 above and accepted or rejected according to the Metropolis-Hastings criterion (9). If crossover is selected, the individuals exchange information as described in 2.1. The pair of offspring is accepted or rejected jointly by using the product of the posterior probabilities of the parents divided by the product of the posterior probabilities of the offspring as the first factor in (9). Note that the crossover operator is defined in such a way that the backward and forward transition probabilities in the second factor of (9) are easily computed.

Note that the EMCMC algorithm is defined in such a way that the stationary distribution consists of independent observations from the posterior distribution of structures given the data. Therefore, the final population after the stopping criterion has been reached can be considered a sample of structures from the posterior distribution. This results in a natural way to obtain valid statistical estimates of properties of the posterior distribution, such as the probability that two nodes are connected by an arc and in which direction, or the probability that a missing observation takes on a particular value.

### 3.3 ADAPTIVE MUTATION

An important characteristic of EMCMC is it uses information from the population of solutions to propose new states. The canonical EMCMC described above does this through the crossover operator. We can also modify the mutation operator to propose new states based on the distribution of states in the current population. The adaptive mutation operator probabilistically proposes changes based on the previous population. This can be thought of as a meta-MCMC, where the overall population is a MCMC. We implemented an adaptive mutation operator for proposing new structures and missing data. Nodes and arcs represent an individual structure in a population. The only difference between the individuals is the placement and direction of the arcs. As the population converges to the stationary distribution, it seems reasonable that the distribution will have more arcs between highly dependent nodes and fewer between conditionally independent nodes. Likewise, the values of the missing data will converge to more likely values given the structure.

To improve acceptance probabilities and thus speed convergence, the proposal distribution for mutation can be a function of the distribution of arc placements in the structure and a distribution of values of missing data. The proposal distribution for mutation of the structures would then by

$$S_S(x, x') = P(\theta_S | I_S, M, \alpha) \quad (11)$$

where $\theta_S$ is the parameter for adding or deleting an arc, $I_S$ is the individual structure currently selected, M is the current population, and $\alpha$ is the distribution of arcs over the population of structures. The proposal distribution of missing values is defined by

$$S_m(y, y') = P(\theta_m | I_m, N, \beta) \quad (12)$$

where $\theta_m$ is parameter for missing values, $I_m$ is the individual missing values, N is the population of missing values, and $\beta$ is the distribution of missing values.

## 4 EMPIRICAL RESULTS

### 4.1 APPROACH

Our approach to evaluating these algorithms was to first find a "good" set of parameters for the algorithms, then compare the algorithms with an appropriate set of parameters. We used a known network, Figure 4, to generate a set of data for training and a separate set of data for test. This allowed us to compare the results to the true model. Each algorithm was run for 500 iterations over 5 repetitions.

The metrics we used were the Bayesian Dirichlet score, the Log loss, best so far curves, and convergence curves. The Bayesian Dirichlet score is the metric used as the fitness function and energy function for each of the algorithms. It is the log posterior probability of the model given the data. In our case, the data consists of the complete data and the imputed missing values. The log loss is a commonly used metric appropriate for probabilistic learning algorithms. It is a member of the family of proper scoring rules. Proper scoring rules have the characteristic that they are maximized when the learned probability distribution corresponds to the empirically observed probabilities. The log loss for a variable X on case i is given by $-\log\{p(x_i)\}$, where $p(x_i)$ is the probability that the Bayesian network assigns, given the values of the observed variable other than X on case i, to the actual observed value $x_i$. We evaluated the algorithms using log loss on a holdout sample.

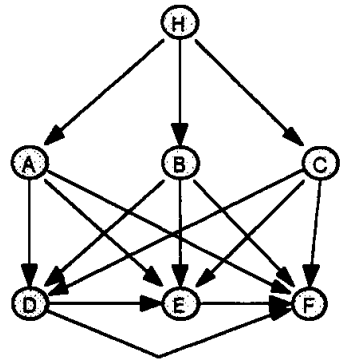

Figure 4: 1X3X3 Network

The best so far curve is used extensively in evolutionary algorithms to measure the trajectory of the best solution overall found at that time. It is a measure of how quickly an algorithm finds the single best solution during that particular run.

The convergence curve is a measure of how fast a Markov chain or population of Markov chains converges to the stationary distribution. We use the convergence metric developed by Gelman and Rubin for measuring convergence of multiple chains [Gelman and Rubin 1992]. The measure uses the within and between chain variances to calculate what they call a scale reduction. A score of 1 indicates convergence. Since a score of 1 is difficult to achieve, Gelman and Rubin recommend using 1.2 or 1.1 to declare convergence. We use both.

### 4.2 ALGORITHM COMPARISONS

After selecting the set of values for the algorithm parameters (crossover rate, mutation rate, etc.) we compared the performance of each algorithm. Figure 5 shows the Bayesian Dirichlet score and log loss for each of the algorithms. The results are shown in 95% credible intervals.

The Bayesian Dirichlet score plots indicate that the EA and MCMC with adaptive mutation find much higher probable networks than the MCMC and EMCMC algorithms. The log loss plots overlap for all the algorithms but there is still an improvement of the EA and MCMC with adaptive mutation over the MCMC and EMCMC. Both the EA and MCMC with adaptive mutation find networks that are more probable given the data than the original networ and networks whose log loss is almost as good as the original network. The Bayesian Dirichlet score for the original network was −3200 while the log loss was 3.10.

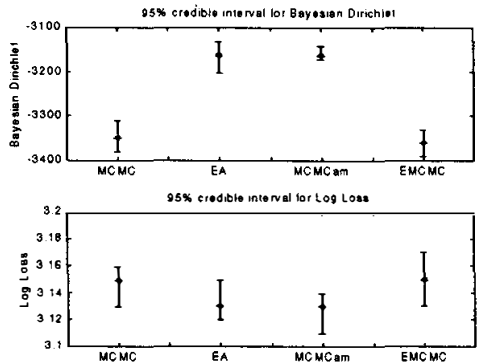

Figure 5: Bayesian Dirichlet score and log loss

The reason the EA performs so well is because it samples from highly probable modes. But the reason for the MCMC with adaptive mutation's superior performance over the MCMC and EMCMC is not obvious. In theory all MCMC algorithms should sample from the stationary distribution after the algorithms have converged.

The reason for choosing a stopping criterion of 500 iterations is because from all indications the MCMC algorithms converged to the stationary distribution prior to 500 iterations. Figure 6 shows the plots of multiple chains and a plot of the Gelman Rubin metric for a single run. Each of the plots led to a reasonable inference that the chains have converged prior to 500 iterations. In actuality, we now know the canonical MCMC and EMCMC algorithms had not converged. Figure 7 depicts multiple chains from a canonical MCMC run and a MCMC with adaptive mutation run. After about 500 iterations the MCMC with adaptive mutation has converged while the canonical MCMC has not converged even after 3000 iterations. In fact, we ran the canonical MCMC over 5000 iterations and the algorithm was still slowly approaching convergence.

These empirical results illustrate two important findings. First, local changes made with global information (i.e. adaptive mutation) can help make dramatic improvements in improving the mixing (speed convergence) of MCMC algorithms. Second, the MCMC algorithm can find equally high probable networks as the EA. If the second conjecture is true, which algorithm is preferred?

The top plot of Figure 8 is a plot comparing the trajectory of the best (highest probable) network found during an EA and MCMC with adaptive mutation run. The trajectories are essentially equal. The bottom plot however, shows the diversity of network structures during the same run. By diversity of structure we mean the number of unique structures at any iteration during the run. The EA quickly homes in on a few "good" structures, in effect sampling from a few highly probable modes. The MCMC on the other hand, maintains a completely diverse population during the entire run. This means the MCMC is exploring the parameter space more efficiently. In addition, as mentioned earlier, we can use samples from the MCMC to make inferences about the parameter space. It is unclear what inferences one can make from samples from the EA. It should be pointed out that more advanced techniques, such as niching, from EA theory may lead to improved performance in terms of population diversity.

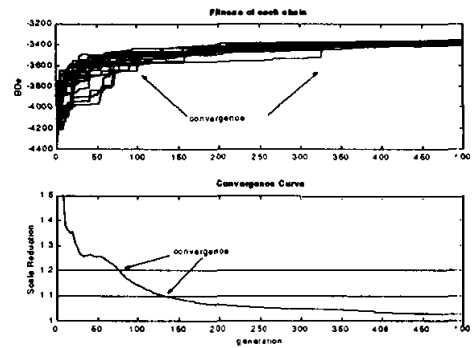

Figure 6: Convergence Plots

## 5 SUMMARY

Learning Bayesian networks from incomplete data is a very difficult problem. The current state-of-the-art approaches use deterministic approaches that get "stuck" at local optima. Our approach is to use stochastic algorithms such as evolutionary algorithms and Markov Chain Monte Carlo algorithms. We also introduce a new hybrid family of algorithms, the Evolutionary Markov Chain Monte Carlo algorithm that combines the benefits of both the EA and MCMC.

We demonstrate that the stochastic algorithms learn Bayesian networks from incomplete data that perform very well. We found that the EA and MCMC algorithms (after convergence) find networks just as probable given the test data as the original network with log loss very close to the original network. Further, we demonstrated that local changes from global information, i.e. adaptive mutation, can make dramatic improvements in MCMC convergence rates.

Further research and empirical tests are needed to explore some of the more advanced techniques from both EA and MCMC. In addition, these methods should be compared to the state-of-the-art greedy algorithms currently in use.

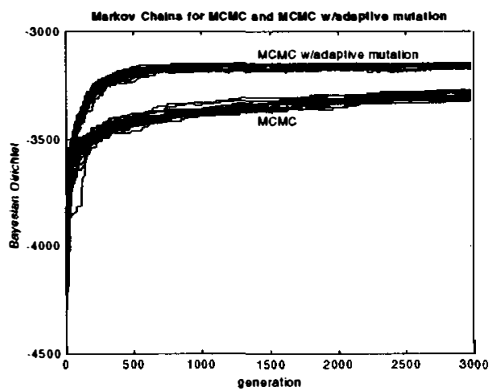

Figure 7: MCMC vs MCMC w/adaptive mutation

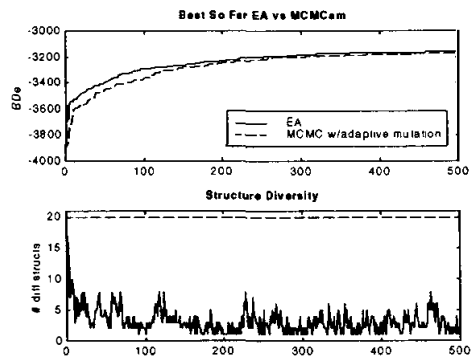

Figure 8: Best So Far and Diversity Plots